\pdfoutput=1

\documentclass[11pt]{article}

\usepackage[final]{acl}

\usepackage{times}
\usepackage{latexsym}
\usepackage{hyperref}
\usepackage{ragged2e}

\usepackage[T1]{fontenc}

\usepackage[utf8]{inputenc}

\usepackage{microtype}

\usepackage{inconsolata}

\usepackage{graphicx}
\usepackage{xcolor}
\usepackage{array}
\usepackage{amsfonts}
\usepackage{amsmath}
\usepackage{booktabs}
\usepackage{xspace}
\usepackage{paralist}
\usepackage{algorithm}
\usepackage{colortbl}
\usepackage{multirow}
\usepackage{cuted}
\usepackage{tcolorbox}
\usepackage{tabularx}
\usepackage{caption}
\usepackage{multicol}
\usepackage{cleveref}
\usepackage{enumitem} 

\newcommand{\modelname}{{ChainRAG}\xspace}
\newcommand{\gptmini}{{GPT4o-mini}\xspace}
\newcommand{\qwen}{{Qwen2.5-72B}\xspace}

\newcommand{\glm}{{GLM-4-Plus}\xspace}
\newcommand{\iterrag}{\textsc{Iter-RetGen}\xspace}

\newcommand{\musique}{MuSiQue\xspace}
\newcommand{\twowiki}{2Wiki\xspace}
\newcommand{\hotpotqa}{HotpotQA\xspace}

\title{Mitigating Lost-in-Retrieval Problems in Retrieval Augmented\\ Multi-Hop Question Answering}

\author{
    Rongzhi Zhu$^\dagger$ \quad 
    Xiangyu Liu$^\dagger$ \quad 
    Zequn Sun$^{\dagger,\,}$\thanks{\,\, Corresponding author} \quad 
    Yiwei Wang$^\ddagger$ \quad
    Wei Hu$^{\dagger,\,\S}$ \\
    $^\dagger$ State Key Laboratory for Novel Software Technology, Nanjing University, China \\
    $^\ddagger$ University of California, Merced, USA \\
    $^\S$ National Institute of Healthcare Data Science, Nanjing University, China \\
    \texttt{\{rzzhu.nju, xyl.nju, wangyw.evan\}@gmail.com, \{sunzq, whu\}@nju.edu.cn} 
}

\begin{document}
\maketitle
\begin{abstract}
In this paper, we identify a critical problem, ``lost-in-retrieval'', in retrieval-augmented multi-hop question answering (QA): the key entities are missed in LLMs' sub-question decomposition. 
``Lost-in-retrieval'' significantly degrades the retrieval performance, which disrupts the reasoning chain and leads to the incorrect answers. 
To resolve this problem, we propose a progressive retrieval and rewriting method, namely ChainRAG, which sequentially handles each sub-question by completing missing key entities and retrieving relevant sentences from a sentence graph for answer generation. Each step in our retrieval and rewriting process builds upon the previous one, creating a seamless chain that leads to accurate retrieval and answers. Finally, all retrieved sentences and sub-question answers are integrated to generate a comprehensive answer to the original question. We evaluate ChainRAG on three multi-hop QA datasets---MuSiQue, 2Wiki, and HotpotQA---using three large language models: GPT4o-mini, Qwen2.5-72B, and GLM-4-Plus. Empirical results demonstrate that ChainRAG consistently outperforms baselines in both effectiveness and efficiency.
\end{abstract}

\section{Introduction}\label{sect:intro}
Large language models (LLMs) \cite{gpt4, glm4, Qwen25, LLMsurvey} have exhibited promising performance on a wide range of natural language processing tasks,
such as machine translation \cite{machine_trasnlation}, text summarization \cite{summary}, question answering (QA) \cite{qa}. 
While LLMs possess strong reasoning abilities, they still face challenges such as outdated knowledge \cite{outdated} and lack of domain-specific expertise \cite{LKM}, which lead to incorrect outputs \cite{Hallucination}.

To fill the gap between LLMs' memory and real-world knowledge, retrieval-augmented generation (RAG) \cite{RAG,ragsurvey,DICL} is widely used to retrieve the knowledge that is relevant to the user's question to improve the LLMs' QA performance. 
Unlike knowledge-enhanced finetuning~\cite{KnowLA}, RAG operates without updating LLM parameters.
When answering multi-hop questions, RAG-based methods generally utilize a question decomposition strategy to decompose the input question into multiple simpler sub-questions.
However, in this decomposition process, we find that when a sub-question lacks a clear entity and instead uses demonstrative pronouns, the retrieval performance drops sharply. 
We refer to this phenomenon as ``\textit{lost-in-retrieval}''. 

\begin{figure}[t]
  \includegraphics[width=\linewidth]{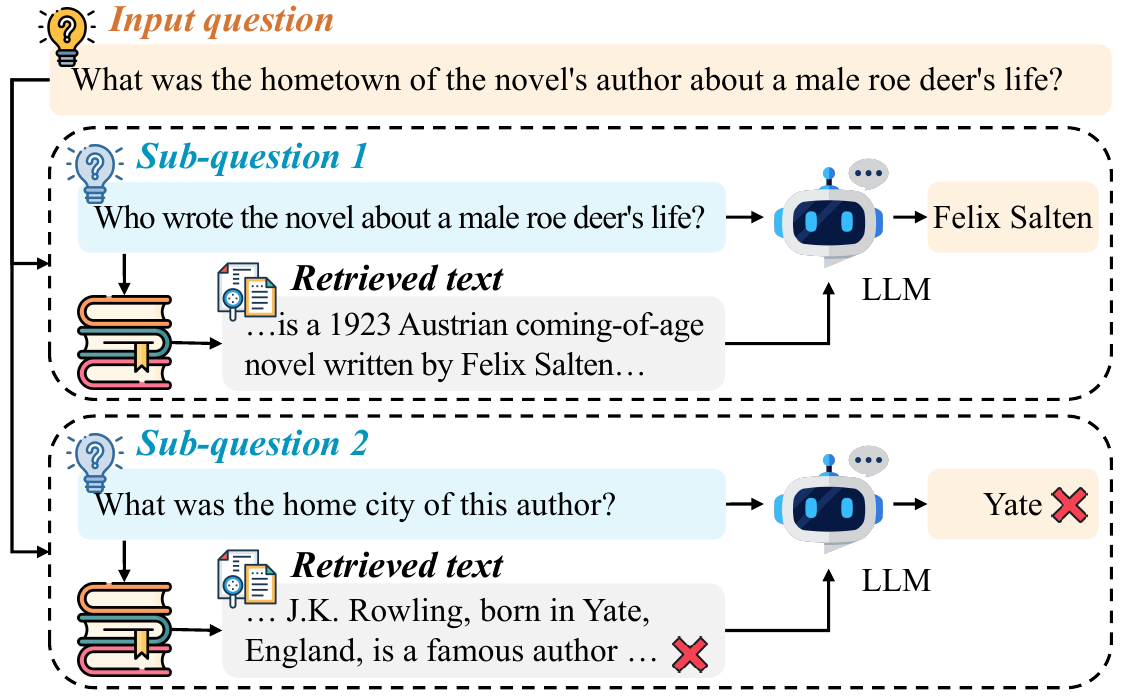}
  \caption{Example of the ``lost in retrieval'' issue where the second sub-question retrieves irrelevant text due to the unclear key entity, leading to an incorrect answer.}
  \label{fig:example}
\end{figure}

\Cref{fig:example} presents a real-world example in solving a multi-hop question using RAG combined with question decomposition.
In this example, the second sub-question, ``What was the home city of this author?'', lacks a clear entity of the author. 
As a result, it leads to retrieval errors, which ultimately cause the final answer to be incorrect.
The right answer is Vienna. 
To give the correct answer, we need to identify the specific key entities in the sub-questions to improve the retrieval performance. 

\begin{figure}[t]
  \includegraphics[width=\linewidth]{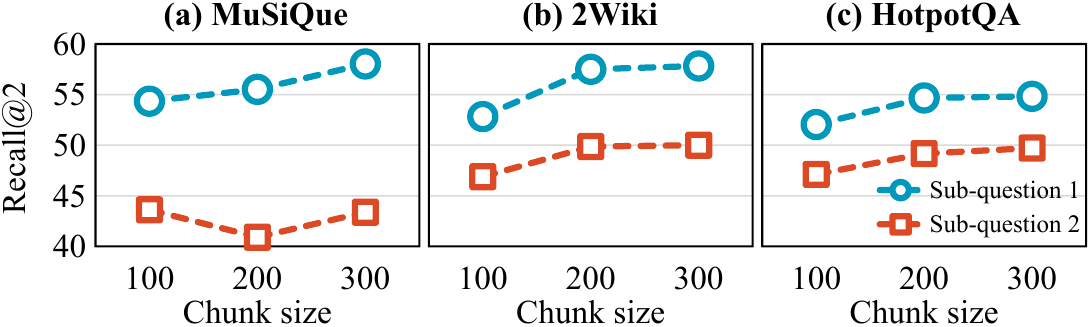}
  \caption{Analysis of ``lost in retrieval''. We evaluate the Recall@2 (\%) of different sub-questions.}
  \label{fig:analysis}
\end{figure}

To analyze the retrieval performance of different sub-questions, we have conducted an empirical study with 300 randomly sampled QA examples from each of the three datasets: \musique~\cite{musique}, 2WikiMultiHopQA (\twowiki) \cite{2wiki}, and \hotpotqa \cite{hotpotqa}. 
Since most questions are two-hop reasoning problems, we calculate the Recall@2 scores for the first two sub-questions.  
As shown in Figure \ref{fig:analysis}, under different chunk size settings, the Recall@2 of the second sub-question is noticeably lower than that of the first sub-question, with an average decrease of 18.29\% across the three datasets.
We analyze the results and find that the first sub-question typically contains a specific key entity, whereas the second sub-question often lacks one.
\textit{The ambiguity of key entities in sub-questions causes the ``lost-in-retrieval'' problems, which further disrupts the chain of reasoning for multi-hop QA.}

To mitigate the ``lost-in-retrieval'' problems and improve multi-hop QA, 
we propose a progressive retrieval framework called \modelname.
It involves an iterative process of sentence retrieval, sub-question answering and subsequent sub-question rewriting. 
We first construct a sentence graph with named entity indexing from texts, which is used to facilitate entity completion in sub-question rewriting and to structure the knowledge scattered across different texts.
Next, given an input question, we employ the LLM to decompose it into several sub-questions and retrieve relevant sentences for the first sub-question.
Then, our iterative process operates as follows until all sub-questions are addressed.
We prompt an LLM to answer the current sub-question.
The answer is then used to rewrite the next sub-question by completing any missing key entities, if possible.
The updated sub-question is subsequently used for retrieval.
Finally, all retrieved sentences and sub-question answers are integrated to answer the original question.

We conduct a series of experiments using three LLMs on three multi-hop QA datasets from LongBench \cite{longbench}, evaluating the performance and efficiency of our method. The results suggest that our method consistently outperforms the baselines across the three datasets. It also demonstrates stable performance across different LLMs, reflecting a certain degree of robustness. In summary, our contributions are outlined as follows:

\begin{itemize}
\item We investigate the ``lost-in-retrieval'' problems of RAG for multi-hop QA. We identify that the reason is the absence of key entities in sub-questions by empirical studies.

\item To resolve this issue, we propose \modelname, a progressive retrieval and sub-question rewriting framework. We construct a sentence graph based on the similarities and entities within the texts to support our retrieval and the completion of missing entities in sub-questions.

\item We evaluate our \modelname  on three multi-hop QA datasets. Our experimental results and analysis show that it outperforms the baselines in both effectiveness and efficiency.
\end{itemize}

\section{Related Work}
In this section, we review related work and discuss how our method differs from them.

\subsection{Retrieval-Augmented Generation}
RAG~\cite{RAG} is a widely-used technique for addressing knowledge-intensive tasks. 
It enables LLMs to fetch relevant information from external knowledge bases, enhancing their effectiveness for complex QA, such as multi-hop KBQA.
The biggest challenge of RAG lies in how to retrieve relevant and comprehensive information.
One solution is to perform multiple rounds of retrieval to gather relevant passages~\cite{IRCOT,IterRAG23}.
The other solution seeks to remove irrelevant information from retrieved texts~\cite{LongRAG_Jiang}. 
Besides, recent work pays more attention to the effective utilization of retrieved texts using techniques like
gist memory~\cite{GIST},
text summarization~\cite{ReCOMP}, 
reranking~\cite{R2G,CoRAG}, 
context compression~\cite{flexrag}.

Despite the aforementioned methods, RAG still encounters challenges when handling long texts or complex questions, since relevant information is usually scattered across different parts of the text.
To address this issue, many studies incorporate graph structures to organize the text. 
RAPTOR~\cite{raptor} structures the text into a tree structure.
GraphRAG~\cite{graphrag}, GraphReader~\cite{graphreader}, and HippoRAG~\cite{HippoRAG} use LLMs to extract entities and relations from the text, constructing knowledge graphs (KGs). 
While effective, these methods rely on LLMs for entities and facts extraction, which increases costs.

Our work investigates a subtle problem in RAG for multi-hop QA, i.e., ``lost in retrieval'', caused by the missing topic entities in sub-questions.
Our method resolves the problem by iteratively completing the missing entities and retrieving relevant sentences containing these entities.
It eliminates the need for a complex reasoning process or an expensive KG construction pipeline.

\subsection{Multi-hop QA}
Multi-hop QA is an ideal scenario for evaluating RAG systems, since it requires strong capabilities from both the knowledge retriever and the answer generator. 
RoG~\cite{RoG} adopts a planning-retrieval-reasoning paradigm, using relation paths in KGs to guide the retrieval of effective reasoning paths.
EfficientRAG~\cite{EfficientRag} fine-tunes the DeBERTa-v3-large model~\cite{Deberta} to construct a labeler and a filter for handling multi-round queries, reducing the frequency of LLM calls.
OneGen~\cite{ONEGen} unifies generation and retrieval by fine-tuning the model to perform both tasks simultaneously in a single-step inference. 
Many existing multi-hop QA methods use LLMs for query decomposition~\cite{ragsurvey}. 
However, as demonstrated in the empirical study in Section~\ref{sect:intro}, this strategy suffers from ``lost-in-retrieval''.
Our work resolves this issue without fine-tuning and frequent API calls.

\begin{figure*}[!t]
  \includegraphics[width=\textwidth]{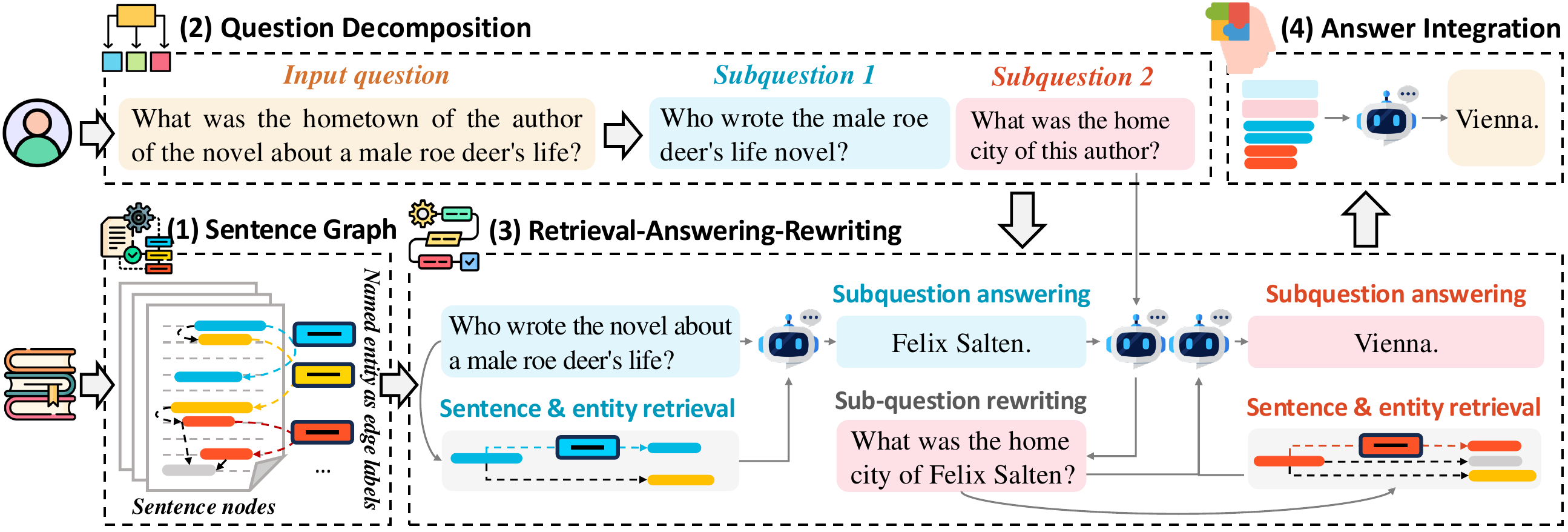}
  \caption{Framework overview of \modelname. 
  It first constructs a sentence graph, where the edges between sentence nodes are labeled by their common named entities.
  Given a question, it is decomposed into sub-questions. 
  Then, our iterative process involves retrieval, answering, and rewriting the unclear sub-question by filling in missing entities. 
  Finally, it integrates all retrieved sentences and answers to produce a comprehensive answer.}
  \label{fig:framework}
\end{figure*}

\section{Methodology}\label{sect:method}
Figure~\ref{fig:framework} provides an overview of \modelname. 
We first construct a sentence graph from texts. 
Given a question, we use an LLM to decompose it into several sub-questions. 
Then, we design an iterative process includes sentence retrieval, sub-question answering, and subsequent sub-question rewriting. 
This process continues until all sub-questions are addressed. Finally, we integrates all retrieved sentences and sub-question answers to produce a comprehensive answer to the original question. 

\subsection{Sentence Graph with Entity Indexing}
In our method, the completion of missing key entities in sub-questions is a crucial step.
To facilitate entity completion, it is essential to identify all named entities retrievable from the given texts.
Therefore, for efficiency consideration, we first extract named entities from texts $s_i$ using spaCy, resulting in an entity set $\mathcal{E}_i$. 
In this process, we also store the mappings between each entity and all its sentences.
Then, to conveniently obtain all the information of an entity from scattered texts for the following knowledge retrieval, we propose to construct a sentence graph with named entities as edge labels, where each node represents a sentence and each edge between nodes indicates that the two sentences describe the same entity.
We denote the node set as \( C = \{s_1, s_2, \dots, s_n\} \), which has already been obtained in the previous entity extract process. 
As for the edge, the entity-sentence mappings can be used to mine edges.
However, relying solely on entity co-occurrence edges is insufficient for effective knowledge retrieval. Besides, the sentence-level retrieval is too fine-grained.
We should enhance the associations between sentences for border and comprehensive retrieval.
Finally, we consider the following three types of edges in our sentence graph:
\begin{itemize}
    \item \textbf{Entity co-occurrence (EC)}. If two sentences describe the same key entity, they will be linked.
    A sentence may contain multiple entities, but not all of them are the key entities. We calculate the importance score, i.e., BM25, for each entity $e \in \mathcal{E}_i$, and retain only the top-$\alpha$\% entities as key entities, denoted by $\mathcal{K}_i \subseteq \mathcal{E}_i$. This process reduces redundancy of the following construction steps. Two sentences $s_i$ and $s_j$ would be linked with an edge labeled ``EC'' if $\mathcal{K}_i \cap \mathcal{K}_j \neq \emptyset$ holds.
    
    \item \textbf{Semantic similarity (SS)}. If two sentences have a high embedding similarity, they will be linked. We encode each sentence $s_i$ into a dense vector $\mathbf{v}_i$ using OpenAI text-embedding-3-small embeddings for computing pairwise similarities of sentences. For sentence $s_i$, we maintain a set $\mathcal{R}_i$ containing its top-$m$ most similar sentences. 
     Two sentences $s_i$ and $s_j$ would be linked with an edge labeled ``SS'' if $s_j \in \mathcal{R}_i \ \lor \ s_i \in \mathcal{R}_j$ holds.
     
    \item \textbf{Structural adjacency (SA)}. If two sentences are adjacent in texts, they will be linked. In this work, we consider a span of three sentences. If two sentences $s_i$ and $s_j$ are within three sentences of each other, i.e., $|i - j| \leq 3$, we add an edge labeled ``SA'' between them. This type of edges can helps us reconstruct the overall structure of text for a wider retrieval.
\end{itemize}

The sentence graph plays a crucial role in mitigating the ``lost-in-retrieval'' problems.
It connects sentences through shared entities and semantic associations, organizing the knowledge to ensure that even when a sub-question lacks clear entities, the necessary context can still be retrieved.

\subsection{Sentence and Entity Retrieval}
Before the retrieval process begins, we first utilize LLM to decompose multi-hop questions into sub-questions. The detailed prompt used for decomposing multi-hop questions can be found in Appendix \ref{appx:prompt}. Our retrieval method deals with the sub-questions in turn,
which is a progressive retrieval process and constructs a complete inference chain through entity expansion. 
It involves the following two retrieval steps for each sub-question.

\paragraph{Seed sentence retrieval.} 
Given a sub-question, we first calculate its embedding similarity (e.g., inner product) with all sentences in the sentence graph.
This can be done quickly by matrix multiplication. 
We then filter out the sentences with low similarity to narrow down the retrieval candidates.
Next, we use a cross-encoder to assess the relevance of each candidate sentence within the context of the sub-question, and finally, we select the top-$k$ sentences as seed sentences.

\paragraph{Retrieval expansion on sentence graph.} 
Starting from seed sentences, we iteratively explore their neighbors in the sentence graph.
After each expansion, we use a LLM to assess whether the gathered sentences contain sufficient information to answer the question. 
If validated, the expansion is terminated. 
Otherwise, we continue to explore higher-order neighbors.
To ensure both efficiency and quality of the retrieval, we implement multiple optimization mechanisms. 
To reduce the number of LLM calls, the initial neighbor exploration fetches all 1-hop neighbors of the seed sentences. 
Additionally, to prevent the context from becoming too lengthy, we introduce a length limit. Once the total length of the retrieved sentences reaches this limit, the retrieval process is stopped. The LLM is finally promoted to answer the sub-question based on the retrieved sentences.

\subsection{Sub-question Rewriting}
As we have mentioned and validated in Sect.~\ref{sect:intro} that the retrieval performance degrades if sub-questions lack necessary named entities.
To resolve this problem, we propose to rewrite sub-questions.
First, we determine if a sub-question needs rewriting by checking for the presence of pronouns (such as ``this'', ``it'', ``they'', etc.). 
If these pronouns exist, we feed both the current sub-question and the previous sub-questions along with their answers into an LLM to rewrite the current sub-question.
When the previous sub-questions were not answered, we cannot rewrite the sub-questions. In this case, we summarize their corresponding context and incorporate the summary into the context of the current sub-question. 
Our sub-question rewriting method serves two purposes.
First, it mitigates the degradation in retrieval performance caused by missing entities. 
Second, since the previous sub-question is closely related to the current one, preserving its key information supports the reasoning process of the current sub-question.

\subsection{Answer and Context Integration}
After obtaining the answers to all sub-questions, we have two different integration methods to generate a comprehensive answer to the original question: sub-answer integration and sub-context integration.

\paragraph{Sub-answer integration.}
This method generates the answer to the original question by utilizing each sub-question and its answer.
It relies solely on the information from the sub-questions' answers, without external interference. Since the decomposition of sub-questions can be regarded as a reasoning process, this method enables the LLM to infer the original question's answer. Moreover, this method ensures that the LLM processes the relevant text of only one sub-question at a time, avoiding performance degradation caused by the LLM's weak long-context processing capabilities when processing multiple sub-questions simultaneously.
Consequently, this method does not require the LLM strong long-context processing capabilities, as the final synthesis relies exclusively on the concise information from the sub-questions and their generated answers, without utilizing any retrieved relevant text.
However, if a sub-question is answered incorrectly or left unanswered, it can significantly impact the final result.
To reduce such impact, we consider the following method that integrates all retrieved sentences to enrich the context.

\paragraph{Sub-context integration.} 
We remove duplicate contexts from all retrieved sentences (without using sub-questions)  and use a cross-encoder to rerank the sentences for generating the final answer. 
This method is similar to traditional RAG, which uses only relevant text to generate the answer. 
It helps mitigate the impact of errors in sub-question decomposition or answers by focusing on the retrieved sentences, rather than relying solely on the sub-question answers.
However, this method explicitly requires the LLM to have strong long-context processing capabilities to effectively process the merged and potentially extensive context. Compared to answering each sub-question individually, the merged context may contain more noisy information and negatively impact the final answer.

\begin{table*}[t]
\small\centering
\resizebox{.99\linewidth}{!}{
\begin{tabular}{llcccccc}
\toprule
\multirow{2}{*}{\textbf{LLMs}} & \multirow{2}{*}{\textbf{Methods}}
& \multicolumn{2}{c}{\textbf{MuSiQue}} & \multicolumn{2}{c}{\textbf{2Wiki}} & \multicolumn{2}{c}{\textbf{HotpotQA}} \\ 
\cmidrule(lr){3-4} \cmidrule(lr){5-6} \cmidrule(lr){7-8}
& & F1 & EM & F1 & EM & F1 & EM \\ 
\midrule
\multirow{7}{*}{\textbf{\gptmini}}
& NaiveRAG & 29.82 & 19.00 & 50.61 & 42.50 & 56.92 & 42.00 \\
& NaiveRAG w/ QD & 37.49 & 26.00 & 56.88 & 38.50 & 60.00 & 43.50\\
\cmidrule(lr){2-8}
& \iterrag Iter3~\cite{IterRAG23}  & 38.41 & 33.00 & 58.43 & \underline{50.50} & 57.77 & 42.00 \\
& LongRAG~\cite{LongRAG_Jiang} & 44.88 & 32.00 & \underline{62.39} & 49.00 & \textbf{64.74} & \textbf{51.00} \\
& HippoRAG w/ IRCoT~\cite{HippoRAG} & 46.50 & 28.50 & 62.38 & 48.00 & 56.12 & 40.00 \\
\cmidrule(lr){2-8}
& {\cellcolor{gray!20}Ours (AnsInt)} & {\cellcolor{gray!20}\textbf{50.54}} & {\cellcolor{gray!20}\underline{37.00}} & {\cellcolor{gray!20}\textbf{62.55}} & {\cellcolor{gray!20}\textbf{52.00}} & {\cellcolor{gray!20}60.73} & {\cellcolor{gray!20}46.00} \\
& {\cellcolor{gray!20}Ours (CxtInt)} & {\cellcolor{gray!20}\underline{47.87}} & {\cellcolor{gray!20}\textbf{38.50}} & {\cellcolor{gray!20}56.54} & {\cellcolor{gray!20}\underline{50.50}} & {\cellcolor{gray!20}\underline{64.59}} & {\cellcolor{gray!20}\underline{50.00}} \\
\midrule
\multirow{7}{*}{\cellcolor{white}{\textbf{\qwen}}}
& NaiveRAG & 27.08 & 16.50 & 39.82 & 28.50 & 50.25 & 34.50 \\
& NaiveRAG w/ QD & 33.91 & 20.50 & 53.84 & 37.00 & 52.14 & 34.50 \\
\cmidrule(lr){2-8}
& \iterrag Iter3~\cite{IterRAG23} & 40.15 & 31.50 & 53.59 & 41.50 & 58.41 & 45.00 \\
& LongRAG~\cite{LongRAG_Jiang} & 40.89 & 29.50 & 62.00 & 51.50 & 60.29 & \underline{46.50} \\
& HippoRAG w/ IRCoT~\cite{HippoRAG} & 44.64 & 31.50 & 64.19 & 52.00 & 55.21 & 41.00\\
\cmidrule(lr){2-8}
& {\cellcolor{gray!20}Ours (AnsInt)} & {\cellcolor{gray!20}\underline{47.75}} & {\cellcolor{gray!20}\underline{37.00}} & {\cellcolor{gray!20}\underline{64.23}} & {\cellcolor{gray!20}\underline{54.00}} & {\cellcolor{gray!20} \underline{60.55}} & {\cellcolor{gray!20}46.00} \\
& {\cellcolor{gray!20}Ours (CxtInt)} & {\cellcolor{gray!20}\textbf{49.37}} & {\cellcolor{gray!20}\textbf{39.00}} & {\cellcolor{gray!20}\textbf{65.85}} & {\cellcolor{gray!20}\textbf{55.50}} & {\cellcolor{gray!20}\textbf{64.54}} & {\cellcolor{gray!20}\textbf{52.00}} \\
\midrule
\multirow{7}{*}{\textbf{\glm}}
& NaiveRAG & 37.86 &28.00 & 57.78 & 45.50 & 58.42 & 44.00 \\
& NaiveRAG w/ QD &  30.33 &22.50 & 60.93 & 48.00 & 59.05 & 44.00\\
\cmidrule(lr){2-8}
& \iterrag Iter3~\cite{IterRAG23} &  \textbf{52.57} & \textbf{44.50} & 66.56 & 56.50 & 61.03 & \underline{48.50}\\
& LongRAG~\cite{LongRAG_Jiang} &  40.44 & 29.00 & 62.27 & 54.50 & 61.60 & \underline{48.50} \\
& HippoRAG w/ IRCoT~\cite{HippoRAG}& 44.70 & 29.50 & \underline{67.55} & 55.50 & \underline{63.91} & 48.00\\
\cmidrule(lr){2-8}
& {\cellcolor{gray!20}Ours (AnsInt)} & {\cellcolor{gray!20}\underline{51.66}} & {\cellcolor{gray!20}\underline{40.00}} & {\cellcolor{gray!20}67.35} & {\cellcolor{gray!20}\underline{58.00}} & {\cellcolor{gray!20}55.40} & {\cellcolor{gray!20}42.00} \\
& {\cellcolor{gray!20}Ours (CxtInt)} & {\cellcolor{gray!20}49.40} & {\cellcolor{gray!20}38.00} & {\cellcolor{gray!20}\textbf{70.58}} & {\cellcolor{gray!20}\textbf{61.50}} & {\cellcolor{gray!20}\textbf{64.22}} & {\cellcolor{gray!20}\textbf{50.00}} \\
\bottomrule
\end{tabular}}
\caption{Performance (\%) on MuSiQue, 2Wiki, and HotpotQA. QD refers to question decomposition. AnsInt refers to generating answers using only sub-questions and their corresponding answers, while CxtInt refers to generating answers using only the contexts retrieved by sub-questions.}
\label{tab:main_results}
\end{table*}

\section{Experiments}
In this section, we report the experimental results and analysis to evaluate the effectiveness and efficiency of our method for multi-hop QA. 
Our source code is available at GitHub.\footnote{\url{https://github.com/nju-websoft/ChainRAG}}

\subsection{Setup}

\paragraph{Dataset and metrics.}
We use the following three challenging multi-hop QA datasets in our experiments: 
\musique~\cite{musique}, \twowiki ~\cite{2wiki}, and \hotpotqa~\cite{hotpotqa}. 
Instead of using raw data, we follow the same data setting as in  LongBench~\cite{longbench}. 
Detailed statistics of the used datasets are provided in the Appendix~\ref{appx:data}.
Following convention, we assess the multi-hop QA performance using the F1-score and exact match (EM) score.

\paragraph{Baselines.}
To ensure the fairness of our evaluation, we standardize the embedding model, i.e., OpenAI's text-embedding-small-v3,\footnote{\url{https://platform.openai.com/docs/guides/embeddings}} and the cross-encoder reranker, i.e., BGE-Reranker~\cite{BGE}, across both our method and the baselines.
We conduct experiments with three popular LLMs as the answer generator: \gptmini,\footnote{\url{https://openai.com/index/gpt-4o-mini-advancing-cost-efficient-intelligence/}} \qwen~\cite{Qwen25} and \glm~\cite{glm4}.
For comparison, we select NaiveRAG and three advanced train-free RAG methods as baselines: 
Iter-RetGen~\cite{IterRAG23}, 
LongRAG~\cite{LongRAG_Jiang}, and a combination of HippoRAG~\cite{HippoRAG} with IRCoT~\cite{IRCOT}. 
Except for the main experiments and the efficiency analysis, all subsequent experiments are conducted exclusively on \gptmini, as similar results have been observed with other LLMs.
Our method has two variants, namely \modelname (AnsInt) and \modelname (CxtInt), which use the sub-answer integration and sub-context integration strategies, respectively.

\paragraph{Implementation details.} 
In all experiments, we set the word limit to 3000. 
For sentence graph construction, $\alpha$ is set to 60 for the entity filter and $m$ is set to 10 for selecting the most similar sentences. 
During the seed sentence retrieval phase, the number of candidate sentences selected in the first round is 100, 
with the top-$k$ sentences chosen as seed sentences, where $k=3$ in the main experiment. 
Further implementation details are provided in Appendix \ref{appx:implement}.

\subsection{Main Results}
We hereby provide a detailed comparison and analysis of the overall results shown in Table \ref{tab:main_results}.
In general, our method has performed better compared to baselines.
Compared to NaiveRAG, \modelname achieves significant improvements across all datasets, especially on \musique, where the average F1 score has improved by approximately 60\%. 
When compared to three advanced RAG methods, 
\modelname consistently outperforms them, achieving the highest average performance across all three datasets. This is most pronounced when using \qwen as the LLM, where the average F1 score of CxtInt is 59.92.
This surpasses the second-best method, HippoRAG w/ IRCoT, which has an average F1 score of 54.68, representing a 9.6\% improvement. This advantage demonstrates the effectiveness of our proposed progressive retrieval and query entity completion strategy.

\modelname has also demonstrated stable performance across all three LLMs, reflecting its robustness. We further observe some differences in the results of each LLM. 
For example, when using \gptmini, our AnsInt variant outperformed CxtInt, while the opposite is true for \qwen and \glm. 
This difference may primarily stem from the varying capabilities of these LLMs. 
\gptmini exhibits strong reasoning abilities, while the other two LLMs are better at processing long-context. 
Our two answering strategies can each leverage these two advantages separately.

We also find that adding question decomposition to NaiveRAG sometimes brings only limited improvements. For example, on the \hotpotqa dataset, the average F1 score increases by just 1.87, while the average EM score improves by only 0.5. 
In addition, when using \glm as the LLM, the performance of NaiveRAG on the \musique dataset decreases after incorporating question decomposition. These cases also suggest the presence of the ``lost-in-retrieval'' problems.
In contrast, \modelname consistently benefits from question decomposition with sub-question rewriting, showing its robustness in resolving the ``lost-in-retrieval'' problems.

\subsection{Ablation Study}
\begin{figure}
  \includegraphics[width=.95\linewidth]{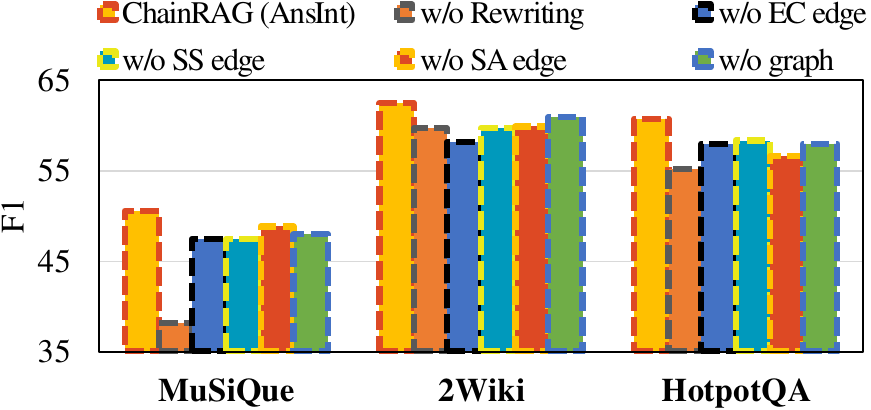}
  \caption{F1 (\%) comparison of ablation study.}
  \label{fig:ablation_f1}
\end{figure}

\begin{figure}
  \includegraphics[width=.95\linewidth]{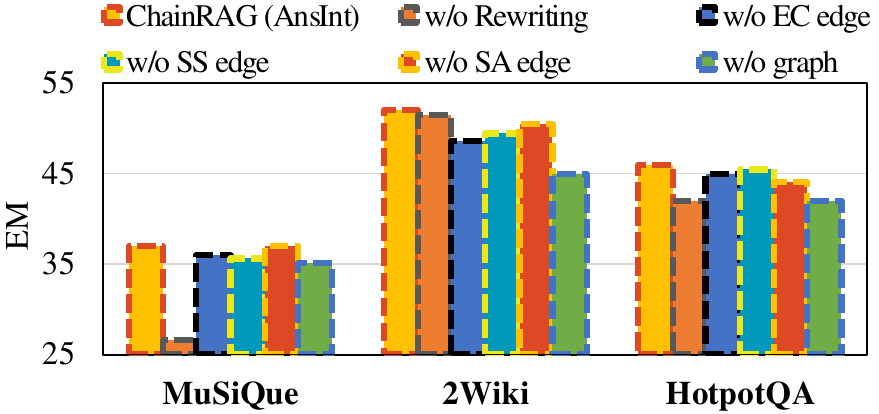}
  \caption{EM (\%) comparison of ablation study.}
  \label{fig:ablation_em}
\end{figure}

To validate the effectiveness of each technical module in our method, we carry out an ablation study. 
We explore the effects of removing the sub-question rewriting method, ablating edges in the sentence graph by type, and completely removing the graph (i.e., segmenting texts into chunks like NaiveRAG). 
Figures \ref{fig:ablation_f1} and \ref{fig:ablation_em} show the F1 and EM scores using the AnsInt strategy, respectively. 
The complete results can be found in Appendix~\ref{appx:ablation}.
We observe similar results when using CxtInt.


Sub-question rewriting is one of the core components of our method. As shown in the experimental results, removing this component leads to a significant decline in QA performance. 
On the \musique dataset, both F1 and EM scores drop by approximately 30\%. 
These results further confirm the existence of the ``lost-in-retrieval'' problems.

Ablating different edge types in our sentence graph leads to performance declines, demonstrating the rationality of our edge design for graph construction. 
Moreover, the impact of ablating specific edge types varies across datasets. For example, removing SA edges has the greatest impact on the \hotpotqa dataset, while the situation is reversed on the \musique dataset. This indicates that different datasets have distinct dependencies on edge types. Future work can dynamically adjust edge construction strategies for each dataset to further improve performance.

To evaluate the effectiveness of our sentence graph, we remove the entire graph and adopt a method similar to NaiveRAG, where we index the texts in chunks.
The results show a decline in both F1 and EM scores, with the decrease in EM being notably more pronounced. We speculate that this is because finer-grained textual information can better guide LLM to generate more accurate answers. This outcome further confirms the rationality and effectiveness of our sentence graph.
 
\begin{table}[t]
\centering
\resizebox{1.0\linewidth}{!}{
    \begin{tabular}{lccc}
    \toprule
    & \textbf{\musique} & \textbf{\twowiki} & \textbf{\hotpotqa} \\
    \midrule
    Sub-question 1 & 55.52  & 57.50  & 54.67  \\
    Sub-question 2 & 40.91  & 49.87  & 49.17  \\
    Modified sub-question 2  & 58.81  & 54.32  & 61.83  \\
    \bottomrule
    \end{tabular}}
    \caption{Recall@2 (\%) results of text retrieval for sub-questions. The modified sub-question 2 is rewritten by our entity completion method.}
  \label{tab:retrieval}
\end{table}%

\subsection{Retrieval Performance Analysis}
To validate whether our method effectively addresses the ``lost-in-retrieval'' problems, we conduct additional retrieval  experiments. Since our method divides the context into sentences, which differs from NaiveRAG in granularity, we standardize the chunk size in \modelname to match NaiveRAG for fair comparison. From Table \ref{tab:retrieval}, we observe that after rewriting, the Recall@2 of sub-question 2 shows a marked improvement. Notably, on the \musique and \hotpotqa datasets, the Recall@2 of the rewritten sub-question 2 even exceeds that of sub-question 1, as the rewriting process uses information from sub-question 1 to retrieve more relevant content.

\begin{figure}[t]
  \includegraphics[width=.95\linewidth]{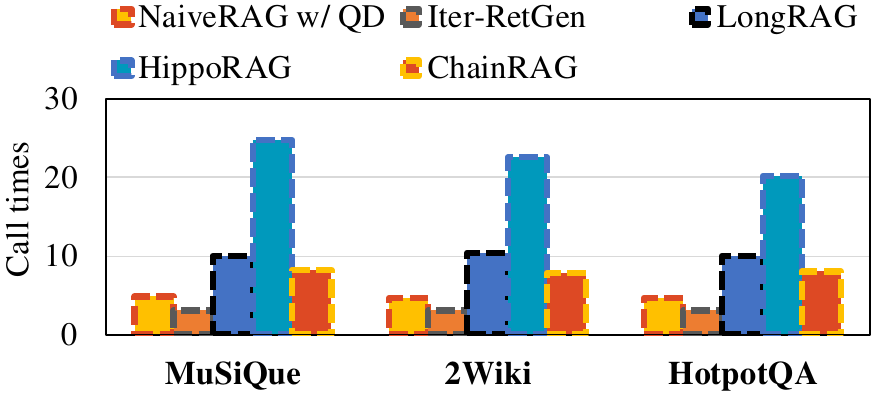}
  \caption{Comparison of \gptmini call times.}
  \label{fig:call}
\end{figure}

\begin{figure}[b]
  \includegraphics[width=\linewidth]{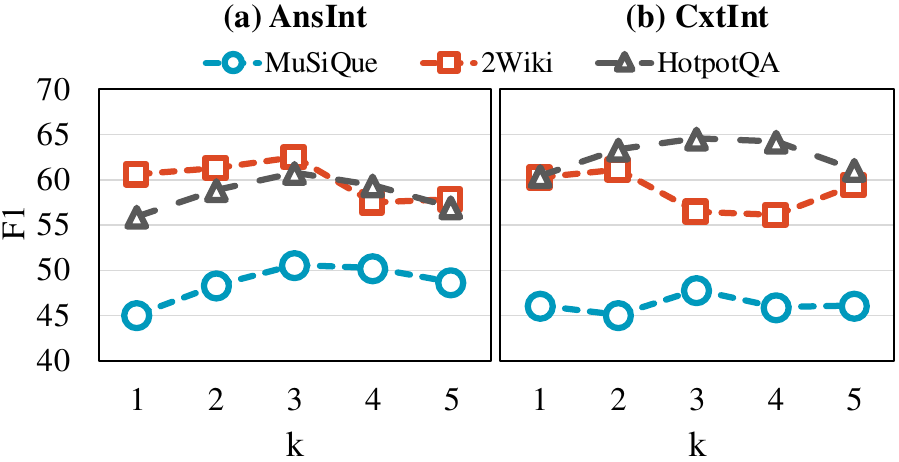}
  \caption{F1 (\%) results w.r.t. different $k$ values.}
  \label{fig:k}
\end{figure}

\begin{table*}[!t]
    \centering
    \footnotesize
    \setlength{\tabcolsep}{4.5pt}
    \begin{tabular}{clllll}\toprule
        & \textbf{Question} & \textbf{Gold facts} & \textbf{Sub-question 1} & \textbf{Sub-question 2} & \textbf{Final answer} \\
        \midrule
        \multirow{2}{*}{\rotatebox[origin=c]{90}{\parbox[c]{3.3cm}{\centering NaiveRAG w/ QD}}}
        & \begin{minipage}[t]{2cm} 
        In what region of the country of S-Fone is The place of birth of John Phan located? 
        \end{minipage}
        & \begin{minipage}[t]{2.6cm} 
        $\Rightarrow$ John Phan born October 10, 1974, in \textcolor{blue}{Da Nang, Vietnam}. \\ 
        $\Rightarrow$ \textcolor{blue}{South Central Coast} (sometimes called South Central Region) consists of the independent municipality of Da Nang. 
        \end{minipage}
        & \begin{minipage}[t]{2.6cm}
        \textbf{Question}: What is the place of birth of John Phan? \\
        \textbf{Context}: John Phan born October 10, 1974, in \textcolor{blue}{Da Nang, Vietnam}. \\ 
        \textbf{Answer}: \textcolor{blue}{Da Nang, Vietnam}.  
        \end{minipage}
        & \begin{minipage}[t]{3.7cm} 
        \textbf{Question}: In which region of S-Fone is this place located?\\context: S-Fone was a mobile communication operator in Vietnam that used the CDMA technology. Founded on 1 July 2003 in \textcolor{red}{Ho Chi Minh City, Vietnam}.\\
        \textbf{Answer}: \textcolor{red}{Ho Chi Minh City, Vietnam}
        \end{minipage}
        & \begin{minipage}[t]{2.0cm} \textcolor{red}{Ho Chi Minh City, Vietnam}. \end{minipage} \\
        \midrule
        \multirow{2}{*}{\rotatebox[origin=c]{90}{\parbox[c]{3.3cm}{\centering \modelname}}}
        & \begin{minipage}[t]{2cm} 
        In what region of the country of S-Fone is The place of birth of John Phan located? \end{minipage}
        & \begin{minipage}[t]{2.6cm}
        $\Rightarrow$ John Phan born October 10, 1974, in \textcolor{blue}{Da Nang, Vietnam}. \\ 
        $\Rightarrow$ \textcolor{blue}{South Central Coast} (sometimes called South Central Region) consists of the independent municipality of Da Nang.
        \end{minipage}
        & \begin{minipage}[t]{2.6cm} \textbf{Question}: What is the place of birth of John Phan?\\
        \textbf{Context}: John Phan born October 10, 1974, in \textcolor{blue}{Da Nang, Vietnam}.\\
        \textbf{Answer}: \textcolor{blue}{Da Nang, Vietnam}.  
        \end{minipage}
        & \begin{minipage}[t]{3.7cm} 
        \textbf{Question}: In which region of S-Fone is Da Nang, Vietnam located?\\
        \textbf{Context}: \textcolor{blue}{South Central Coast} (sometimes called South Central Region) consists of the independent municipality of Da Nang and seven other provinces.\\
        \textbf{Answer}: \textcolor{blue}{South Central Coast}
        \end{minipage}
        & \begin{minipage}[t]{2.0cm} 
        \textcolor{blue}{South Central Coast}. \end{minipage} \\
        \bottomrule

    \end{tabular}
    \caption{Examples of the RAG process of NaiveRAG w/ QD and our \modelname from the \musique dataset. Blue text represents correct and relevant information or answers, while red text indicates incorrect information.}
    \label{tab:nor-oner-cot-examples}
\end{table*}

\subsection{Efficiency Comparison}
We measure the efficiency of each method by counting the number of LLM calls. Figure \ref{fig:call} compares the average number of calls of different methods across three datasets.
HippoRAG, which constructs knowledge graphs based on LLM, requires several times more calls than other methods and is significantly influenced by the dataset. This is especially true for datasets with longer context lengths, where the number of calls increases further. Compared to LongRAG, \modelname shows a notable improvement in efficiency across all datasets, with an average reduction of about 17.3\% in the number of calls. Although \iterrag is the most efficient method (second only to NaiveRAG), considering the overall performance of the three models across the three datasets, \modelname achieves a better balance between efficiency and performance.

\subsection{Top-k Study}
The selection of \( k \) sentences as seed sentences in the retrieval step is crucial, as it significantly influences the subsequent retrieval process. Given the fixed word limit, a smaller \( k \) value tends to retrieve more sentences from higher-order neighbors of the seed sentences, while a larger \( k \) value retrieves more sentences from lower-order neighbors. 
To determine the optimal value of \( k \), we conduct comparative experiments. As shown in Figure \ref{fig:k}, except for CxtInt's performance on the \twowiki dataset, all other cases achieve the best results when \( k=3 \). Considering the overall performance, we choose \( k=3 \) as the default value.

\subsection{Case Study}
We present an example question from \musique and compare the RAG process of NaiveRAG w/ QD and \modelname in Table \ref{tab:nor-oner-cot-examples}. As NaiveRAG w/ QD does not rewrite the sub-question, its second sub-question lacks a clear entity, leading to retrieval errors and causing the ``lost-in-retrieval'' problems. This ultimately results in an incorrect answer. 
In the example, the context comes from the first chunk retrieved, which is also the source of the final answer. 
For \modelname, by using the answer from the first sub-question to complete the entity in the second sub-question, the retrieved information becomes more accurate,
on which the LLM then provides the correct answer.

\subsection{Further Analysis}

Our method consists of several key steps, including entity recognition, question decomposition, and entity recovery.
To assess their impact, we conduct additional experiments using \qwen.

\subsubsection{Impact of Entity Recognition}
Evaluating entity recognition performance on QA datasets is challenging since there is no established gold standard. 
To evaluate the robustness of \modelname in entity recognition, we integrate different entity recognition methods and compare their impact on the final QA performance.
Specifically, we use \qwen for entity recognition and compare the QA F1 scores on \musique against those obtained using spaCy.
The results are presented in Table \ref{tab:ner_comparison}.
We can see that the performance gains achieved with \qwen are modest. 
Our method is robust in entity extraction.
Considering efficiency and effectiveness trade-offs, spaCy remains a well-balanced choice for entity extraction.

\begin{table}[!t]
\centering
\begin{tabular}{lcc}
    \toprule
\textbf{Method} & \textbf{spaCy } & \textbf{\qwen} \\
    \midrule
Ours (AnsInt)   & 47.75      & 49.21            \\
Ours (CxtInt)   & 49.37      & 50.61            \\
    \bottomrule
\end{tabular}
\caption{F1 scores on the MuSiQue dataset using different entity recognition methods with \qwen.}
\label{tab:ner_comparison}
\end{table}

\subsubsection{Impact of Question Decomposition and Entity Recovery}
To evaluate question decomposition and entity recovery, we manually review the results of \qwen on the test examples (600 questions in total) of three datasets. 
On average, 89.3\% of questions are successfully decomposed into reasonable sub-questions. Decomposition errors typically arise from questions containing lengthy modifiers or complex phrasing. 
In contrast, the accuracy of entity recovery is slightly lower, with an average score of 79.3\%. This discrepancy can be attributed to the inherent difficulty of the entity recovery task, which requires LLMs not only to identify the demonstrative pronoun to be replaced but also to integrate the answer from the preceding sub-question to perform the replacement.
Furthermore, to analyze the effect of question decomposition (QD) and entity recovery (ER) on the final QA performance, 
we divide the test examples into the following four groups:

\begin{itemize}
    \item \textbf{Incorrect QD:} The test examples that have incorrect question decomposition.
    \item \textbf{Incorrect ER:} The test examples that have correct question decomposition but incorrect entity recovery results.
    \item \textbf{Others:} The remaining test examples that have correct decomposition and entity recovery, or the examples do not require entity recovery (e.g., parallel multi-hop questions).
    \item \textbf{Total:} The total test examples.
\end{itemize}

We calculate the F1 scores for each group using the sub-context integration method with \qwen. 
Despite incorrect question decomposition, the sub-context integration method demonstrates robustness, achieving an average F1 score of 55.43 by leveraging all retrieved information. 
This is often because one sub-question in an incorrect decomposition still resembles the original question, helping retrieval. 
However, incorrect entity recovery has a more significant negative impact, leading to ``lost-in-retrieval'' problems and lower performance compared to the ``Others'' group, which is unaffected by such problems. Overall, entity recovery errors are more detrimental to performance than question decomposition errors, underscoring the importance of addressing ``lost-in-retrieval'' problems. However, despite these problems, our method consistently outperforms the baselines, demonstrating its overall effectiveness and robustness.

\begin{table}[!t]
\centering
\resizebox{1.0\linewidth}{!}{
\begin{tabular}{lccc}
\toprule
\textbf{} & \textbf{MuSiQue} & \textbf{2Wiki} & \textbf{HotpotQA} \\
\midrule
Incorrect QD & 40.79 & 63.19 & 62.28 \\
Incorrect ER & 39.75 & 57.58 & 56.69 \\
Others & 54.30 & 67.12 & 65.83 \\
\midrule
Total & 49.37 & 65.85 & 64.54 \\
\bottomrule
\end{tabular}}
\caption{F1 scores of different groups \modelname (CxtInt) with Qwen2.5-72B.}
\label{tab:qd_er_impact}
\end{table}

\section{Conclusion and Future Work}\label{sect:concl}
In this paper, we investigate the ``lost-in-retrieval'' problems of RAG that occurs during multi-hop QA. 
We propose a progressive retrieval framework involving sentence graph construction, question decomposition, retrieval, and sub-question rewriting, which effectively enhances the retrieval performance, especially for sub-questions lacking clear entities. 
Our experiments and analysis on three challenging datasets---MuSiQue, 2Wiki, and HotpotQA---demonstrate that our method consistently outperforms baselines. 
Additionally, it demonstrates robustness and efficiency across different LLMs, showcasing its adaptability and potential for broader applications.

For future work,we plan to optimize efficiency by exploring lightweight graph traversal and adaptive termination strategies, reducing LLM calls and resource consumption.
We also plan to enhance dynamic adaptability by developing dataset-specific edge construction policies to better align with diverse text structures.

\section*{Acknowledgments}
This work was funded by National Natural Science Foundation of China (Nos. 62406136 and 62272219), 
Natural Science Foundation of Jiangsu Province (No. BK20241246),
Postdoctoral Fellowship Program of CPSF (No. GZC20240685),
China Postdoctoral Science Foundation (No. 2024M761396),
and Jiangsu Funding Program for Excellent Postdoctoral Talent.

\section*{Limitations}

While our proposed ChainRAG framework has demonstrated significant improvements in resolving the ``lost-in-retrieval'' problems, several limitations should be acknowledged.
First, although ChainRAG outperforms LongRAG and HippoRAG in efficiency, our iterative process of retrieval, sub-question rewriting increases the computational resources and time required compared to NaiveRAG. 
This may pose challenges for deployment in resource-constrained environments.
Second, although robust across existing datasets, adapting ChainRAG to highly specialized domains requires further validation and potential adjustments to indexing strategies.
Third, the accuracy of entity recognition and completion is critical for the success of ChainRAG. Errors in entity recognition can propagate through the retrieval and reasoning process, affecting the overall performance.

\bibliography{custom}

\appendix

\section{Dataset Statistics}\label{appx:data}

Dataset statistics are presented in Table \ref{tab:data_stastics}. To meet the requirements of LongRAG and HippoRAG, we have segmented the context within the dataset. Since these contexts are originally composed of multiple integrated paragraphs, the original segmentation structure of the data can be accurately restored by regular expression matching.

\begin{table}[!h]
\small\centering\setlength{\tabcolsep}{5pt}
    \begin{tabular}{lrrr}
    \toprule
    \textbf{Datasets} & \textbf{\musique} & \textbf{\twowiki} & \textbf{\hotpotqa} \\
    \midrule
    No. of Samples & 200  & 200  & 200  \\
    No. of $p$ & 2,212  & 1,986  & 1,722  \\
    Avg. Length  & 11,214  & 4,887  & 9,151  \\
    \bottomrule
    \end{tabular}
    \caption{Statistics of the datasets used in our work. ``Avg. Length'' denotes the average word count.}
  \label{tab:data_stastics}
\end{table}

\section{Implementation Details}\label{appx:implement}

We strictly follow the process outlined in the Iter-RetGen~\cite{IterRAG23} paper and the prompts provided in its appendix for our reproduction. Additionally, we refer to the results in the paper and the EfficientRAG~\cite{EfficientRag}, selecting the third iteration results as the baseline, as the third iteration often produces results close to the optimal and exhibits a clear edge effect.
For LongRAG and HippoRAG, we use their open-source code, follow the default settings, and conduct testing after switching to the unified model.
Regarding the chunking process, LongRAG follows the default settings and segments based on word count, with a chunk size of 200. HippoRAG, on the other hand, does not apply any chunking. 

We utilize the ``en\_core\_web\_sm'' model from the spaCy library for entity extraction. This model is a small-scale model with approximately millions of parameters, designed for lightweight and efficient natural language processing tasks.

\begin{table}[t]\centering
\resizebox{1.0\linewidth}{!}{
\begin{tabular}{lcccccc}
\toprule
Methods & \musique & \twowiki & \hotpotqa \\
\midrule
\modelname (AnsInt) & 50.54 & 62.55 & 60.73 \\
\phantom{GPT}w/o Rewriting & 38.18 & 59.69 & 55.19 \\
\phantom{GPT}w/o EC edge & 47.52 & 58.28 & 57.99 \\
\phantom{GPT}w/o SS edge & 47.40 & 59.78 & 58.40 \\
\phantom{GPT}w/o SA edge & 48.83 & 59.87 & 56.64 \\
\phantom{GPT}w/o Graph & 47.97 & 60.96 & 58.05 \\
\midrule
\modelname (CxtInt) & 47.87 & 56.54 & 64.59 \\
\phantom{GPT}w/o Rewriting & 36.60 & 56.36 & 61.57 \\
\phantom{GPT}w/o EC edge & 45.34 & 55.12 & 61.73 \\
\phantom{GPT}w/o SS edge & 43.39 & 55.68 & 60.85 \\
\phantom{GPT}w/o SA edge & 47.36 & 55.08 & 61.32 \\
\phantom{GPT}w/o Graph & 40.18 & 54.76 & 60.19 \\
\bottomrule
\end{tabular}}
\caption{\label{tab:ablatioin_f1}F1 results of ablation study.}
\end{table}

\begin{table}[t]\centering
\resizebox{1.0\linewidth}{!}{
\begin{tabular}{lcccccc}
\toprule
Methods & \musique & \twowiki & \hotpotqa \\
\midrule
\modelname (AnsInt) & 37.00 & 52.00 & 46.00 \\
\phantom{GPT}w/o Rewriting & 26.50 & 51.50 & 42.00 \\
\phantom{GPT}w/o EC edge & 36.00 & 48.50 & 45.00 \\
\phantom{GPT}w/o SS edge & 35.50 & 49.50 & 45.50 \\
\phantom{GPT}w/o SA edge & 37.00 & 50.50 & 44.00 \\
\phantom{GPT}w/o Graph & 35.00 & 45.00 & 42.00\\
\midrule
\modelname (CxtInt) & 38.50 & 50.50 & 50.00 \\
\phantom{GPT}w/o Rewriting & 25.00 & 47.00 & 47.00 \\
\phantom{GPT}w/o EC edge & 33.00 & 47.00 & 47.00 \\
\phantom{GPT}w/o SS edge & 34.00 & 46.00 & 48.00 \\
\phantom{GPT}w/o SA edge & 36.50 & 46.50 & 47.00 \\
\phantom{GPT}w/o Graph & 27.00 & 39.50 & 44.00 \\
\bottomrule
\end{tabular}}
\caption{\label{tab:ablatioin_em}EM results of ablation study.}
\end{table}

\begin{figure*}[!t]
\fontsize{10pt}{\baselineskip}\selectfont 
\begin{tcolorbox}
[colback=white,
]

You are a helpful AI assistant that helps break down questions into minimal necessary sub-questions.\\

\textbf{Guidelines}:

1. Only break down the question if it requires finding and connecting multiple distinct pieces of information.

2. Each sub-question should target a specific, essential piece of information.

3. Avoid generating redundant or overlapping sub-questions.

4. For questions about impact/significance, focus on:

\ \ \ \ - What was the thing/event.

\ \ \ \ - What was its impact/significance.

5. For comparison questions between two items (A vs B):

\ \ \ \ - First identify the specific attribute being compared for each item.

\ \ \ \ - Then ask about that attribute for each item separately.

\ \ \ \ - For complex comparisons, add a final question to compare the findings.

6. Please consider the following logical progression:

\ \ \ \ - Parallel: Independent sub-questions that contribute to answering the original question. Example: \{\textit{ex.}\}.

\ \ \ \ - Sequential: Sub-questions that build upon each other step-by-step. Example: \{\textit{ex.}\}.

\ \ \ \ - Comparative: Questions that compare attributes between items. Example: \{\textit{ex.}\}.

7. Keep the total number of sub-questions minimal (usually 2 at most).  Output format should be a JSON array of sub-questions. For example: \{\textit{examples of sub-questions}\}.\\

\textbf{Remember}: 

Each sub-question must be necessary and distinct. Do not create redundant questions. For comparison questions, focus on gathering the specific information needed for the comparison in the most efficient way.

\end{tcolorbox}

\noindent\begin{minipage}{\linewidth}
\captionof{figure}{Prompt for instructing LLMs to decompose the input question into several sub-questions.}\label{fig:prompt1}
\end{minipage}
\end{figure*}

\section{Detailed Results of Ablation Study}\label{appx:ablation}

Tables \ref{tab:ablatioin_f1} and \ref{tab:ablatioin_em} present the complete results of our ablation study. After removing certain processes from ChainRAG, we observe varying degrees of performance degradation in QA tasks.
When the rewriting phase is removed, the performance on \musique and \hotpotqa drops significantly, while the performance on \twowiki dataset sees a minor decrease. This is mainly because many of the questions in the \twowiki dataset, after being decomposed, result in two parallel sub-questions with no dependency, which avoids the ``lost-in-retrieval'' problems. 
For \musique, removing the SS edges has the greatest impact on performance, while for \twowiki, it is the removal of the EC edges that has the most significant effect. In \hotpotqa, the SA edges have the largest impact when removed. 
When we remove the entire graph and implement the indexing process according to NaiveRAG, we observe a noticeable decline in performance, with the CxtInt method being more significantly affected. This indicates the effectiveness of our method, which involves building an index using sentences and entities and performing retrieval on the sentence graph.

\section{Prompt Example}\label{appx:prompt}
Figure \ref{fig:prompt1} shows our prompt to guide LLMs in decomposing complex multi-hop questions into a series of minimal and necessary sub-questions. 
The goal of this decomposition is to ensure that each sub-question targets a specific and essential piece of information, thereby facilitating more accurate and efficient retrieval and reasoning processes. The \textit{ex.} here refers to an example of a question decomposition that follows this logical progression.

\section{Additional Experiment on Small LLMs}
To test the effectiveness of our method on small LLMs, we select two open-source small models, Qwen2.5-14B and Qwen2.5-7B, and compare their performance against the strongest two baselines from the main results.
The results in Table \ref{tab:small_llm} show that our method still outperforms the baselines, showcasing its effectiveness and reliability. These results also reflect the effectiveness of small LLMs in decomposition, rewriting, and summarization.

\begin{table}[!t] 
\centering 
\small
\resizebox{1.0\linewidth}{!}{
\begin{tabular}{llccc}

\toprule
& &\textbf{MuSiQue} & \textbf{2Wiki} & \textbf{HotpotQA} \\
\midrule
\multirow{4}{*}{\rotatebox[origin=c]{90}{\textbf{14B}}} & LongRAG & 36.59 & 54.95 & 58.90 \\
& Hipporag w/ IRCoT     & 37.09  & 58.50 & 54.51  \\
& Ours (AnsInt)  & \textbf{44.47}   & 59.71  & 57.06   \\
& Ours (CxtInt)  & 43.16  & \textbf{63.14} & \textbf{59.43}    \\
\midrule
\multirow{4}{*}{\rotatebox[origin=c]{90}{\textbf{7B}}} & LongRAG  & 29.01 & 49.02& 54.71 \\
& Hipporag w/ IRCoT  & 29.30   & 51.17  & 53.79 \\
& Ours (AnsInt) & 30.90   & 48.25   & 48.50      \\
& Ours (CxtInt)  & \textbf{33.15}   & \textbf{52.18} & \textbf{60.03}    \\
\bottomrule
\end{tabular}}
  \caption{Performance of our \modelname using Qwen2.5-14B and 7B on MuSiQue, 2Wiki, and HotpotQA.} 
  \label{tab:small_llm} 
\end{table}

\begin{table*}[!t] \small
\centering
\begin{tabularx}{\textwidth}{@{} l X @{}}
\toprule
\multicolumn{2}{@{}l@{}}{\textbf{A. Examples of nodes}} \\
\midrule
\textbf{Node ID} & \textbf{Corresponding sentence} \\
\midrule
Node1   & Artificial intelligence has made tremendous progress in recent years. \\
Node2   & Deep learning models can process vast amounts of data and identify patterns. \\
Node3   & These models perform particularly well in image recognition tasks. \\
Node4   & Natural language processing is an important branch of AI. \\
Node5   & NLP enables machines to understand and generate human language. \\
Node6   & This technology has been widely applied in virtual assistants. \\
Node7   & Deep learning algorithms typically require large amounts of data. \\
Node8   & Image recognition can help autonomous vehicles identify road signs. \\
Node9   & Ethical considerations are an important factor in AI development. \\
Node10  & Machine learning models learn from examples rather than explicit programming. \\
Node11  & Neural networks are inspired by the structure of the human brain. \\
Node12  & The development of AI may have significant impacts on the job market. \\
Node13  & Reinforcement learning has enabled AI to master complex games. \\
Node14  & AI systems are increasingly being used in healthcare for diagnosis. \\
Node15  & Privacy concerns arise when AI systems collect and analyze personal data. \\
\midrule[\heavyrulewidth] 
\multicolumn{2}{@{}l@{}}{\textbf{B. Examples of edges}} \\ 
\midrule
\textbf{Edge type} & \textbf{Examples} \\
\midrule
SA edges & (Node1, SA, Node2), (Node2, SA, Node3), (Node3, SA, Node4), (Node4, SA, Node5), \dots \\
\addlinespace
SS edges & (Node1, SS, Node13), (Node1, SS, Node12), (Node4, SS, Node1), (Node6, SS, Node1), (Node9, SS, Node15), \dots \\
\addlinespace
EC edges & (Node2, EC, Node3), (Node2, EC, Node10), (Node3, EC, Node10), (Node2, EC, Node7), \dots \\
\bottomrule
\end{tabularx}
\caption{An example of our sentence graph.}
\label{tab:sentence_graph_example}
\end{table*}

\begin{table*}[!t]
\centering
\small
\begin{tabularx}{\textwidth}{@{} >{\hsize=0.9\hsize}X >{\hsize=1.2\hsize}X >{\hsize=0.6\hsize}X >{\hsize=1.2\hsize}X @{}}
\toprule
\textbf{Original question } & \textbf{Sub-question 1} & \textbf{Rewriting} & \textbf{Sub-question 2} \\
\midrule
\RaggedRight
\textbf{Original question:} What record label is the performer who released \textit{All Your Faded Things} on?

\textbf{ChainRAG answer:} \textcolor{red}{{Blue Note.}}

\textbf{Golden answer:} \textcolor{blue}{Kill Rock Stars.}
&

\textbf{First sub-question:}
Who is the performer of the song \textit{All Your Faded Things}?

\textbf{Retrieved context:}
{Personnel: Anna Oxygen - vocals, piano, composer, art design \dots}

\textbf{LLM answer:} \textcolor{red}{{Unable to answer the question.}}

\textbf{Correct answer:} \textcolor{blue}{Anna Oxygen.}
&

\textbf{Rewriting: }
Cannot rewrite, as the previous sub-question answer is ``unable to answer''.
&

\textbf{Second sub-question:} 
What record label is associated with this performer?

\textbf{Retrieved context:}
Jack Wilson featuring performances recorded and released on the Blue Note label\dots

\textbf{LLM answer:}\enspace \textcolor{red}{{Blue Note.}}

\textbf{Correct answer:}\enspace \textcolor{blue}{Kill Rock Stars.}
\\
\bottomrule
\end{tabularx}
\caption{A failure case study.}
\label{tab:failure_case}
\end{table*}

\section{An Example of Sentence Graph}
To help explain our sentence graph more clearly, we build an example using 15 sentences and present it in Table~\ref{tab:sentence_graph_example}.
SA edges connect directly adjacent sentences like (Node1, SA, Node2), (Node2, SA, Node3).
For SS edges, such as (Node1, SS, Node13) and (Node1, SS, Node12), these connections indicate that Node13 and Node12 are the two sentences most semantically similar to Node1 within this graph.
Other SS edges like (Node4, SS, Node1) and (Node6, SS, Node1) further illustrate these similarity-based connections.
EC edges illustrate how sentences sharing key entities are linked. 
For instance, Node2, Node3, and Node10 are interconnected, forming edges like (Node2, EC, Node3), (Node2, EC, Node10), and (Node3, EC, Node10), 
because they all share the key entity ``models''. 
Similarly, an EC edge exists between Node2 and Node7 since they both contain the key entity ``data''.

\section{Failure Case Analysis}

In this analysis, we provide a real error example from our test results to gain deeper insights into potential deficiencies that may constrain our method.
From this example detailed in Table~\ref{tab:failure_case}, we can observe several key issues. 
For the first sub-question, although the retrieved context contained the correct answer (``Anna Oxygen''), the LLM failed to provide it and instead responded with ``Unable to answer''. We attribute this to potential limitations in the LLM's ability to process long contexts effectively.
This failure meant that the required entity for the second sub-question was unavailable, leading to a ``lost-in-retrieval'' problem.  Consequently, the retrieval process for the second sub-question failed to obtain relevant contextual information about ``Anna Oxygen''. This chain of errors ultimately led to the incorrect answer (``Blue Note'').

\end{document}